\pdfoutput=1
\documentclass[runningheads]{llncs}
\usepackage{graphicx}
\usepackage{amsmath,amssymb} 
\usepackage{color}
\usepackage{cite}
\usepackage[colorlinks = true, citecolor=blue]{hyperref}

\newcommand{\yty}[1]{\textcolor{black}{#1}}
\newcommand{\ytyy}[1]{\textcolor{black}{#1}}
\newcommand{\abc}[1]{\textcolor{black}{#1}}
\newcommand{\abcn}[1]{\textcolor{black}{#1}}

\newcommand{\ty}[1]{\textcolor{black}{#1}}
\newcommand{\tyy}[1]{\textcolor{black}{#1}}

\begin{document}
\title{Learning Dynamic Memory Networks for Object Tracking} 

\titlerunning{Learning Dynamic Memory Networks for Object Tracking}
%
\author{Tianyu Yang \and Antoni B. Chan}
%
\authorrunning{T. Yang and A.B. Chan}
%

\institute{Department of Computer Science, City University of Hong Kong, Hong Kong, China
\email{tianyyang8-c@my.cityu.edu.hk, abchan@cityu.edu.hk}}
\maketitle              
\begin{abstract}
	Template-matching methods for visual tracking have gained popularity recently due to their comparable performance and fast speed. However, they lack effective ways to adapt to changes in the target object's appearance, making their tracking accuracy still far from state-of-the-art. In this paper, we propose a dynamic memory network to adapt the template to the target's appearance variations during tracking. An LSTM is used as a memory controller, where the input is the search feature map and the outputs are the control signals for the reading and writing process of the memory block. As the location of the target is at first unknown in the search feature map, an attention mechanism is applied to concentrate the LSTM input on the potential target. To prevent aggressive model adaptivity, we apply gated residual template learning to control the amount of retrieved memory that is used to combine with the initial template. Unlike tracking-by-detection methods where the object's information is maintained by the weight parameters of neural networks, which requires expensive online fine-tuning to be adaptable, our tracker runs completely feed-forward and adapts to the target's appearance changes by updating the external memory. Moreover, \tyy{unlike other tracking methods where the model capacity is fixed after offline training} -- the capacity of our tracker can be easily enlarged as the memory requirements of a task increase, which is favorable for memorizing long-term object information. Extensive experiments on OTB and VOT demonstrates that our tracker MemTrack performs favorably against state-of-the-art tracking methods while retaining real-time speed of 50 fps. \footnote{Code is available at \url{https://github.com/skyoung/MemTrack}}
	
\end{abstract}

\section{Introduction}

Along with the success of convolution neural networks in object recognition and detection,  an increasing number of trackers \cite{Song2017, Nam2016, Wang2015, Bertinetto2016, Guo2017} have adopted deep learning models for visual object tracking. Among them are two dominant tracking strategies. One is the {\em tracking-by-detection} scheme that online trains an object appearance classifier \cite{Song2017, Nam2016} to distinguish the target from the background. The model is first learned using the initial frame, and then fine-tuned using the training samples generated in the subsequent frames based on the newly predicted bounding box. The other scheme is {\em template matching}, which adopts either the target patch in the first frame \cite{Bertinetto2016, Tao2016} or the previous frame \cite{Held2016} to construct the matching model. To handle changes in target appearance, 
the template built in the first frame may be interpolated by the recently generated object template with a small learning rate \cite{Valmadre2017}.  

The main difference between these two strategies is that tracking-by-detection maintains the target's appearance information in the weights of the deep neural network, thus requiring online fine-tuning with stochastic gradient descent (SGD) to make the model adaptable,
while in contrast, template matching stores the target's appearance in the object template, which is generated by a feed forward computation.  Due to the computationally expensive model updating required in tracking-by-detection, the speed of such methods are usually slow, e.g.\ 
\cite{Song2017, Nam2016, Nam2016-1} run at about 1 fps,
although they do achieve state-of-the-art tracking accuracy. 
%
Template matching methods, however, are fast 
because there is no need to update the parameters of the neural networks. Recently, several trackers \cite{Bertinetto2016, Guo2017, Yang2017} adopt fully convolutional Siamese networks as the matching model, which demonstrate promising results and real-time speed.  However, there is still a large performance gap between template-matching models and tracking-by-detection, due to the lack of an effective method for adapting to appearance variations online.

In this paper, we propose a dynamic memory network, where the target information is stored and recalled from  external memory,  to maintain the variations of object appearance for template-matching.
Unlike tracking-by-detection  where the target's information is stored in the weights of neural networks and therefore \ty{the capacity of the model is fixed after offline training}, the model capacity of our memory networks can be easily enlarged by increasing the size of external memory, which is useful for memorizing long-term appearance variations. 
Since aggressive template updating is prone to overfit recent frames and the initial template is the most reliable one,
we use the initial template as a conservative reference of the object and a residual template, 
obtained from retrieved memory, to adapt to the appearance variations.
During tracking, the residual template is 
\abc{gated channel-wise and} 
combined with the initial template to form the final matching template, which is then convolved with the search image features to get the response map.
\abc{The channel-wise gating of the residual template controls how much each channel of the retrieved template should be added to the initial template, which can be interpreted as a feature/part selector for adapting the template.}
An LSTM (Long Short-Term Memory) is used to control the reading and writing process of external memory, 
\abc{as well as the channel-wise gate vector for the residual template.}
%
In addition, as the target position is at first unknown in the search image, we adopt an attention mechanism to locate the object roughly 
in the search image, thus leading to a soft representation of the target for the input to the LSTM controller. This helps to retrieve the most-related template in the memory. 
%
The whole framework is differentiable and therefore can be trained end-to-end with SGD. In summary, the contributions of our work are:
\begin{itemize}
	\item We design a dynamic memory network for visual tracking. An external memory block, which is controlled by an LSTM with attention mechanism, allows adaptation to appearance variations. 
	\item We propose \abc{gated} residual template learning to generate the final matching template, which effectively controls the amount of appearance variations in retrieved memory that is added to \abc{each channel of} the initial matching template.
	This prevents excessive model updating, while retaining the conservative information of the target.
	\item We extensively evaluate our algorithm on large scale datasets OTB and VOT. Our tracker performs favorably against state-of-the-art tracking methods while possessing real-time speed of 50 fps.
\end{itemize}

\section{Related Work}
\textbf{Template-Matching Trackers}. Matching-based methods have recently gained popularity due to its fast speed and comparable performance. The most notable is the fully convolutional Siamese networks (SiamFC) \cite{Bertinetto2016}. Although it only uses the first frame as the template, SiamFC achieves competitive results and fast speed. The key deficiency of SiamFC is that it lacks an effective model for online updating. 
To address this, \cite{Valmadre2017} proposes model updating using linear interpolation of new templates with a small learning rate, but does only sees modest improvements in accuracy.
Recently, the RFL (Recurrent Filter Learning) tracker \cite{Yang2017} adopts a convolutional LSTM for model updating, where the forget and input gates control the linear combination of historical target information, \emph{i.e.}, memory states of LSTM, and incoming object's template automatically. Guo \emph{et al.} \cite{Guo2017} propose a dynamic Siamese network with two general transformations for target appearance variation and background suppression.
To further improve the speed of SiamFC, \cite{Huang2017} 
reduces the feature computation cost for easy frames, by using deep reinforcement learning to train policies for early stopping the feed-forward calculations of the CNN when the response confidence is high enough.
%
SINT \cite{Tao2016} also uses Siamese networks for visual tracking and has higher accuracy, but runs much slower than SiamFC (2 fps vs 86 fps) due to the use of deeper CNN (VGG16) for feature extraction, and optical flow for its candidate sampling strategy. Unlike other template-matching models that use sliding windows or random sampling to generate candidate image patches for testing, GOTURN \cite{Held2016} directly regresses the coordinates of the target's bounding box by comparing the previous and current image patches. Despite its advantage on handling scale and aspect ratio changes and fast speed, its tracking accuracy is much lower than other state-of-the-art trackers. 

Different from existing matching-based trackers where the capacity of adaptivity is limited by the size of neural networks, we use  SiamFC \cite{Bertinetto2016} as the baseline feature extractor and extend it to use an addressable memory,  whose memory size is independent of neural networks and thus can be easily enlarged as memory requirements of a task increase, to adapt to variations of object appearance.

\textbf{Memory Networks}. Recent use of convolutional LSTM for visual tracking \cite{Yang2017} shows that memory states 
are useful for object template management over long timescales. Memory networks are typically used to solve simple logical reasoning problem in natural language processing like question answering and sentiment analysis. The pioneering works include NTM (Neural Turing Machine) \cite{Graves2014} and MemNN (Memory Neural Networks) \cite{Weston2015}. They both propose an addressable external memory with reading and writing mechanism -- NTM focuses on problems of sorting, copying and recall, while MemNN aims at language and reasoning task. MemN2N 
\cite{Sukhbaatar2015} further improves MemNN by removing the supervision of supporting facts, which makes it trainable in an end-to-end fashion. Based on their predecessor NTM, 
\cite{Graves2016} proposes a new framework called DNC (Differentiable Neural Computer), which uses a different access mechanism to alleviate the memory overlap and interference problem.
Recently, NTM is also applied to one-shot learning \cite{Santoro2016} by redesigning the method for reading and writing memory, and has shown promising results at 
encoding and retrieving new information quickly. 

Our proposed memory model differs from the aforementioned memory networks in the following aspects. Firstly, for question answering problem, the input of each time step is a sentence,
\emph{i.e.}, a sequence of feature vectors (each word corresponds to one vector) which needs an embedding layer (usually RNN) to obtain an internal state. While for object tracking, the input is a search image which needs a feature extraction process (usually CNN) to get a more abstract representation. Furthermore, for object tracking, the target's position in the search image patch is unknown, and here we propose an attention mechanism to highlight the target's information when generating the read key for memory retrieval. 
Secondly, the dimension of feature vector stored in memory for natural language processing is relatively small (50 in MemN2N vs 6$\times$6$\times$256=9216 in our case). 
Directly using the original template for address calculation is time-consuming.  Therefore we apply an average pooling on the feature map to generate a template key for addressing, which is efficient and effective experimentally. 
Furthermore, we apply \abc{channel-wise gated} residual template learning for model updating, and redesign the memory writing operation 
to be more suitable for visual tracking.

\section{Dynamic Memory Networks for Tracking}

\begin{figure*}[t]
	\begin{center}
		\includegraphics[width=0.95\linewidth]{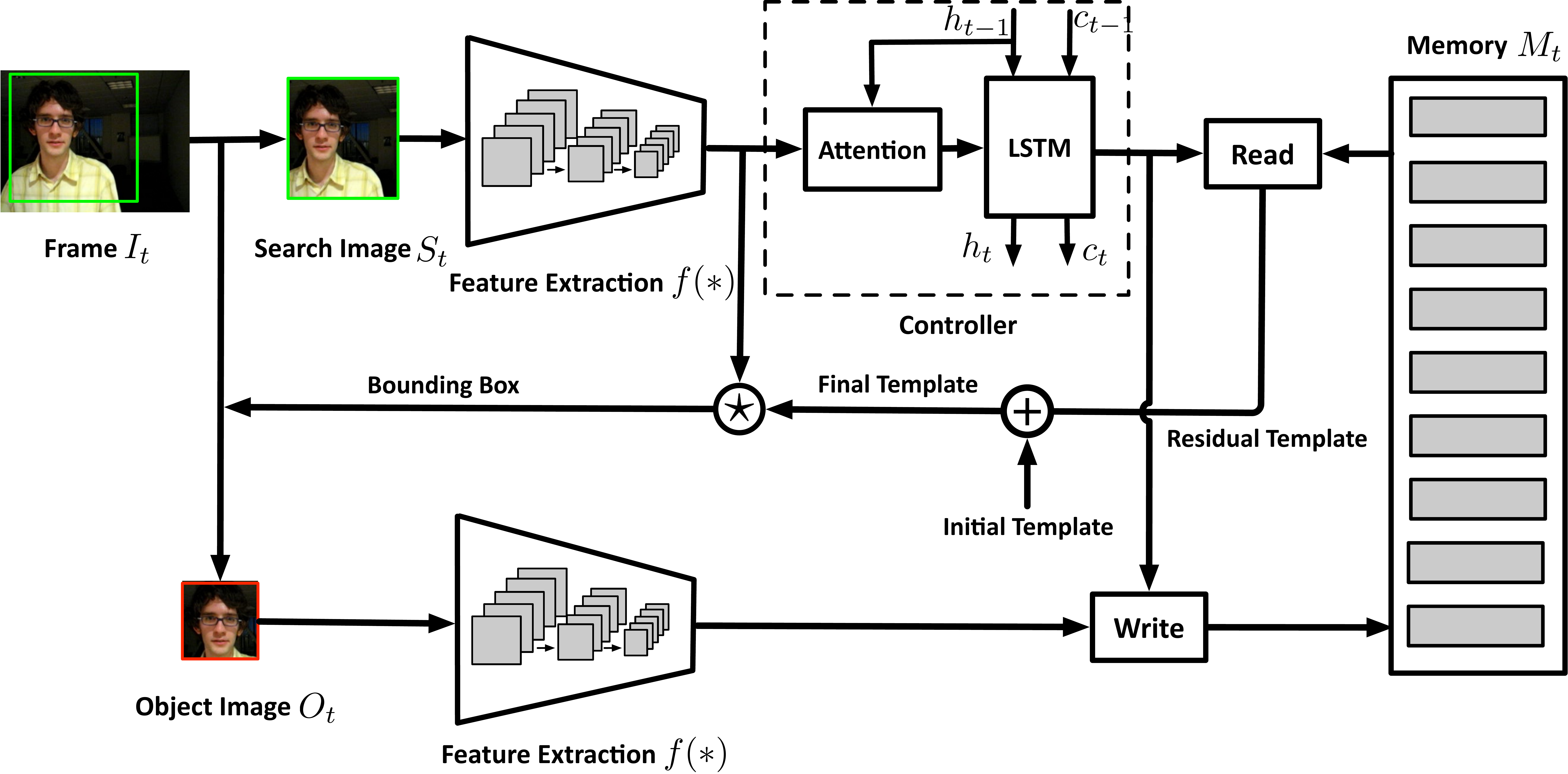}
	\end{center}
	\caption{The pipeline of our tracking algorithm. The green rectangle are the candidate region for target searching. The \textit{Feature Extractions} for object image and search image share the same architecture and parameters. An attentional LSTM extracts the target's information on the search feature map, which guides the memory reading process to retrieve a matching template.  The residual  template is combined with the initial template, to obtain a final template for generating the response score. The newly predicted bounding box is then used to crop the object's image patch for memory writing. 
	}
	\label{fig:2}
\end{figure*}

In this section we propose a dynamic memory network with reading and writing mechanisms for visual tracking. 
The whole framework is shown in Figure \ref{fig:2}.
Given the search image, first features are extracted with a CNN.
The image features are input into an attentional LSTM, which controls the memory reading and writing. 
A residual templates is read from the memory and combined with the initial template learned from the first frame, forming the final template.  The final template is convolved with the search image features to obtain the response map, and the target bounding box is predicted.
The new target's template is cropped using the predicted bounding box, features are extracted and then written into memory for model updating. 

\subsection{Feature Extraction}

Given an input image $I_t$ at time $t$, we first crop the frame into a search image patch $S_t$ with a rectangle that is computed by the previous predicted bounding box. Then it is encoded into a high level representation $f(S_t)$, which is a spatial feature map, via a fully convolutional neural networks (FCNN).  In this work we use the FCNN structure from SiamFC \cite{Bertinetto2016}. 
After getting the predicted bounding box, we use the same feature extractor to compute the new object template for memory writing.

%

\subsection{Attention Scheme}

Since the object information in the search image is needed to retrieve the related template for matching, but the object location is unknown at first, we apply an attention mechanism to make the input of LSTM concentrate more on the target.
We define $\mathbf{f}_{t,i} \in \mathbb{R}^{n \times n \times c}$ as the $i$-th $\mathit{n\times n\times c}$ square patch on $f(S_t)$ in a sliding window fashion.\footnote{We use $6\times6\times256$, which is the same size of the matching template.}
Each square patch covers a certain part of the search image. An attention-based weighted sum of these square patches can be regarded as a soft representation of the object, which can then be fed into LSTM to generate a proper read key for memory retrieval. However the size of this soft feature map is still too large to directly feed into LSTM. 
To further reduce the size of each square patch, 
we first adopt an average pooling with $n\times n$ filter size on $f(S_t)$,
\begin{align}
f^*(S_t) = \text{AvgPooling}_{n\times n}(f(S_t))
\end{align}
and $\mathbf{f}^*_{t,i} \in \mathbb{R}^{c}$ is the feature vector 
for the $i$th patch. 

The attended feature vector is then computed as the weighted sum of the feature vectors,
\begin{align}
\mathbf{a}_t = \sum_{i=1}^{L}\alpha_{t,i}\mathbf{f}^*_{t,i}
\end{align}
where $L$ is the number of square patches, and the attention weights $\alpha_{t,i}$ is calculated by a softmax, 
\begin{align}
\alpha_{t,i} = \frac{\exp(r_{t,i})}{\sum_{k=1}^{L}\exp(r_{t,k})}
\end{align}
where 
\begin{align}
r_{t,i} = W^a \text{tanh}(W^h \mathbf{h}_{t-1}+W^f \mathbf{f}^*_{t,i}+b)
\end{align}
is an attention network which takes the previous hidden state $\mathbf{h}_{t-1}$ of the LSTM controller and a square patch $\mathbf{f}^*_{t,i}$ as input. $W^a, W^h, W^f$ and $b$ are weight matrices and biases for the network.

\begin{figure}[t]
	\begin{center}
		\includegraphics[width=0.9\linewidth]{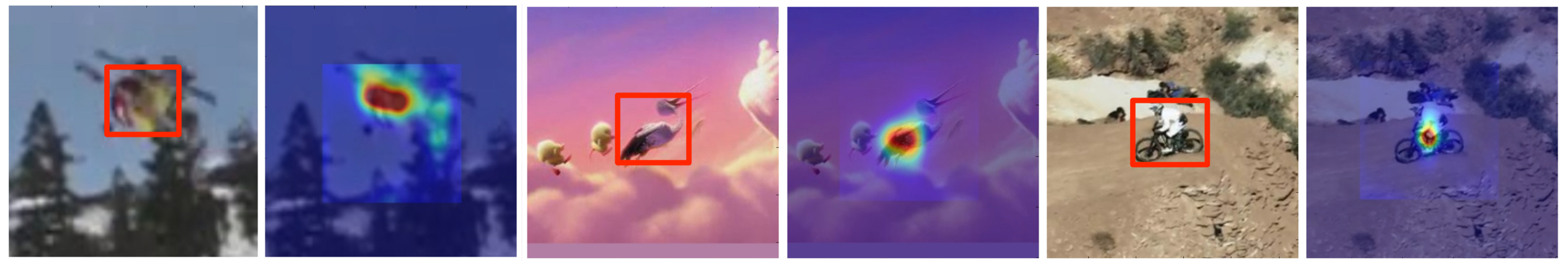}
	\end{center}
	\caption{Visualization of attentional weights map: \abcn{for each pair, (left) search images and ground-truth target box, and (right) attention maps over search image.}
		For visualization, the attention maps are resized using bicubic interpolation to match the size of the original image.}
	\label{fig:3}
\end{figure}

By comparing the target's historical information in the previous hidden state with each square patch, the attention network can generate attentional weights that have higher values on the target and smaller values for surrounding regions.  Figure \ref{fig:3} shows example search images with attention weight maps. We can see that our attention network can always focus on the target which is beneficial when retrieving memory for template matching. 

\subsection{LSTM Memory Controller}

For each time step, the LSTM controller takes the attended feature vector $\mathbf{a}_t$, obtained in the attention module, and the previous hidden state $\mathbf{h}_{t-1}$ as input, and outputs the new hidden state $\mathbf{h}_t$ to calculate the memory control signals, including read key, read strength, bias gates, and decay rate (discussed later).
\abcn{The internal architecture of the LSTM uses the standard model (details in the Supplemental), while the output layer is modified to generate the control signals.}
In addition, we also use layer normalization \cite{Ba2016} and dropout regularization \cite{Srivastava2014} for the LSTM. The initial hidden state $\mathbf{h}_0$ and cell state $\mathbf{c}_0$  
are  
obtained by passing the initial target's feature map through one $n\times n$ average pooling layer and two separate fully-connected layer with tanh activation functions, respectively.

\subsection{Memory Reading}
\begin{figure}[t]
	\begin{center}
		\includegraphics[width=0.61\linewidth]{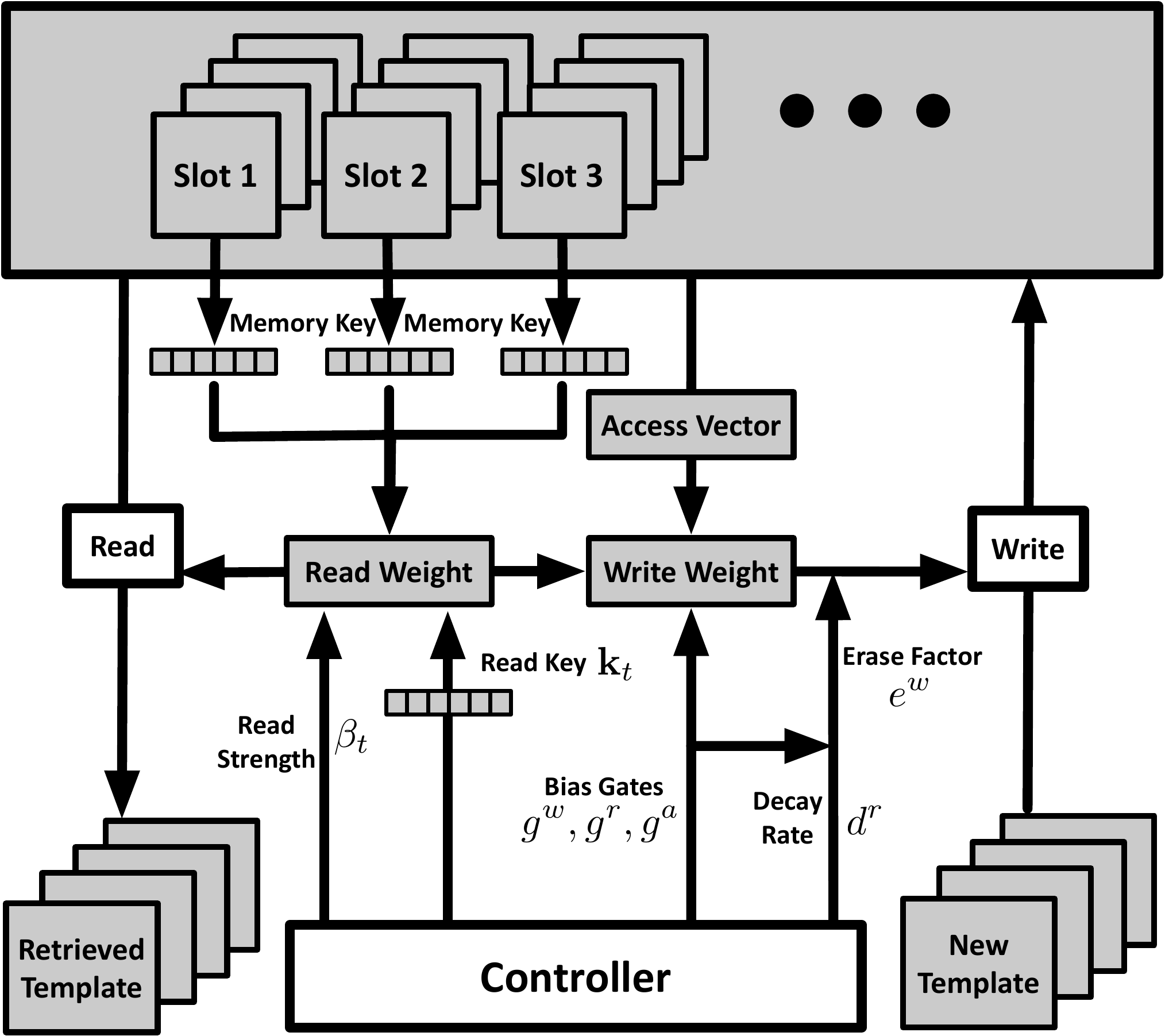}
	\end{center}
	\caption{Diagram of memory access mechanism.}
	\label{fig:4}
\end{figure}

Memory is retrieved by computing a weighted summation of all memory slots with a read weight vector, which is determined by the cosine similarity between a read key and the memory keys. This aims at retrieving the most related template stored in memory.
Suppose $\mathbf{M}_t \in \mathbb{R}^{N\times n \times n \times c}$ represents the memory module,  such that $\mathbf{M}_t(j) \in \mathbb{R}^{n \times n \times c}$ is the template stored in the $j\text{th}$ memory slot and $N$ is the number of memory slots. 
The LSTM controller outputs the read key $\mathbf{k}_t \in \mathbb{R}^{c}$ and read strength $\beta_t \in [1,\infty]$,
\begin{align}
\mathbf{k}_t = & W^k\mathbf{h}_{t}+b^k \\
\beta_t = & 1+\log(1+\exp(W^\beta \mathbf{h}_{t}+b^\beta))
\end{align}
where 
$W^k, W^\beta, b^k, b^\beta$ are corresponding weight matrices and biases.
The read key $\mathbf{k}_t$ is used for matching the contents in the memory, while the read strength $\beta_t$ indicates the reliability of the generated read key. 
Given the read key and read strength, a \textit{read weight} $\mathbf{w}^r_t\in \mathbb{R}^{N}$ is computed for memory retrieval,
\begin{align}
\mathbf{w}^r_t(j) =\frac{\exp{\{C(\mathbf{k}_t, \mathbf{k}_{\mathbf{M}_t(j)})}\beta_t\}}{\sum_{j'} \exp{\{C(\mathbf{k}_t, \mathbf{k}_{\mathbf{M}_t(j')})}\beta_t\}}
\end{align}
where $\mathbf{k}_{\mathbf{M}_t(j)} \in \mathbb{R}^{c}$ is the memory key generated by a $n\times n$ average pooling on $\mathbf{M}_t(j)$. $C(\mathbf{x}, \mathbf{y})$ is the  cosine similarity between vectors, 
$C(\mathbf{x},\mathbf{y})= \frac{\mathbf{x} \cdot \mathbf{y}}{\|\mathbf{x}\|\|\mathbf{y}\|}$.
Finally, the template is retrieved from memory as a weighted sum,
\begin{align}
\mathbf{T}^{\text{retr}}_t=\sum_{j=1}^N\mathbf{w}^r_t(j)\mathbf{M}_t(j).
\end{align}

\subsection{Residual Template Learning}

Directly using the retrieved template for similarity matching  is prone to overfit recent frames.
Instead, we learn a residual template by multiplying the retrieved template with a channel-wise gate vector and add it to the initial template to capture the appearance changes. Therefore, our final template is formulated as,
\begin{align}
\mathbf{T}^{\text{final}}_t = \mathbf{T}_0+ \mathbf{r}_t\odot \mathbf{T}^{\text{retr}}_t,
\end{align}
where $\mathbf{T}_0$ is the initial template and  $\odot$ is channel-wise multiplication.
$\mathbf{r}_t\in \mathbb{R}^c$ is the \textit{residual gate} produced by LSTM controller, 
\begin{align}
\mathbf{r}_t = \sigma (W^r\mathbf{h}_{t}+b^r),
\end{align}
where $W^r, b^r$ are corresponding weights and biases, and $\sigma$ represents sigmoid function. 
The \textit{residual gate} controls how much each channel of the retrieved template is added to the initial one, which can be regarded as a form of feature selection. 

\begin{figure}[t]
	\begin{center}
		\includegraphics[width=0.65\linewidth]{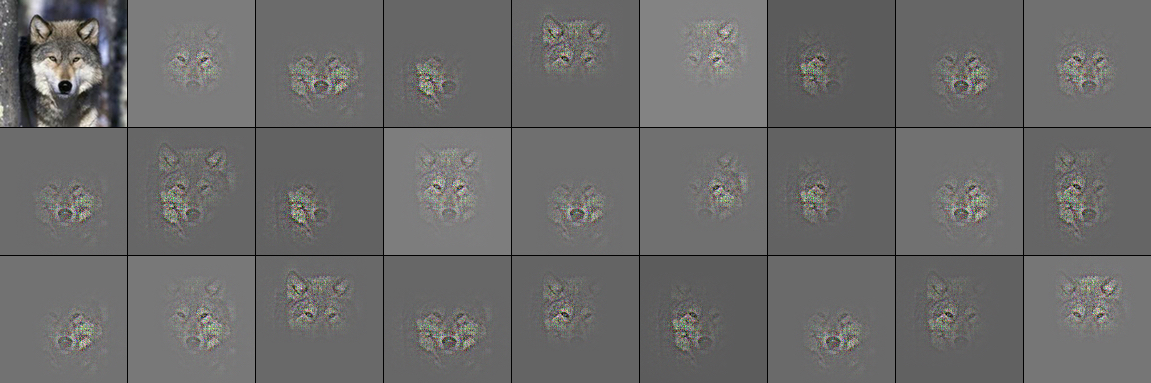}
	\end{center}
	\caption{The feature channels respond to target parts: images are reconstructed from conv5 of the CNN used in our tracker. Each image is generated by accumulating reconstructed pixels from the same channel. The input image is shown in the top-left. }
	\label{fig:6}
\end{figure}

By projecting different channels of a target feature map to pixel-space using deconvolution, as in \cite{Zeiler2014}, we find that the channels focus on different object parts (see Figure \ref{fig:6}). 
Thus, the channel-wise feature residual learning has the advantage of updating different object parts separately. Experiments in Section \ref{abla} show that this yields a big performance improvement. 


\subsection{Memory Writing}

The image patch with the new position of the target is used for model updating, \emph{i.e.}, memory writing.
The new object template $\mathbf{T}^{\text{new}}_t$ is computed using the feature extraction CNN. There are three cases for memory writing: 1) when the new object template is not reliable (e.g.\ contains a lot of background), there is no need to write new information into memory; 2) when the new object appearance does not change much compared with the previous frame, the memory slot that was previously read should be updated; 
3) when the new target has a large appearance change, a new memory slot should be overwritten.
To handle these three cases, we define the \textit{write weight} as
\begin{align}
\mathbf{w}^w_t =g^w\mathbf{0}+g^r\mathbf{w}^r_t + g^a\mathbf{w}^a_t, 
\end{align}
where $\mathbf{0}$ is the zero vector, $\mathbf{w}^r_t$ is the read weight, and $\mathbf{w}^a_t$  is the allocation weight, which is responsible for allocating a new position for memory writing. 
The write gate $g^w$, read gate $g^r$ and allocation gate $g^a$, are produced by the LSTM controller with a softmax function, 
\begin{align}
[g^w, g^r, g^a] = \text{softmax}(W^g \mathbf{h}_{t}+b^g),
\end{align}
where $W^g, b^g$ are the weights and biases. Since $g^w+g^r+g^a=1$, these three gates govern the interpolation between the three cases.  If $g^w=1$, then $\mathbf{w}^w_t=\mathbf{0}$ and nothing is written.  If $g^r$ or $g^a$ have higher value, then the new template is either used to update the old template (using $\mathbf{w}^r_t$) or written into newly allocated position (using $\mathbf{w}^a_t$). The \textit{allocation weight} is calculated by,
\begin{align}
\mathbf{w}^a_t(j)=
\begin{cases}
1, &\text{if } j=\displaystyle \mathop{\mathrm{argmin}}_{j} \mathbf{w}^u_{t-1}(j)\\
0, &\text{otherwise}
\end{cases}
\end{align}
where $\mathbf{w}^u_t$ is the \textit{access vector},
\begin{align}
\mathbf{w}^u_t = \lambda \mathbf{w}^u_{t-1} + \mathbf{w}^r_t + \mathbf{w}^w_t,
\end{align}
which indicates the frequency of memory access (both reading and writing), and $\lambda$ is a decay factor. Memory slots that are accessed infrequently will be assigned new templates.  

The writing process is performed with a \textit{write weight} in conjunction with an \textit{erase factor} for clearing the memory, 
\begin{align}
\mathbf{M}_{t+1}(j) = \mathbf{M}_{t}(j)(\mathbf{1}-\mathbf{w}^w_t(j)e^w)+\mathbf{w}_t(j)^we^w\mathbf{T}^{\text{new}}_t,
\end{align}
where 
$e^w$ is the \textit{erase factor} computed by
\begin{align}
e^w = d^rg^r+g^a,
\end{align}
and $d^r \in [0,1]$ is the \textit{decay rate} produced by the LSTM controller, 
\begin{align}
d^r = \sigma (W^d\mathbf{h}_{t}+b^d),
\end{align}
where $\sigma$ is sigmoid function. $W^d$ and $b^d$ are corresponding weights and biases. If $g^r=1$ (and thus $g^a=0$), then $d^r$ serves as the decay rate for updating the template in the memory slot (case 2). If $g^a=1$ (and $g^r=0$), $d^r$ has no effect on $e^w$, and thus the memory slot will be erased before writing the new template (case 3). Figure \ref{fig:4} shows the detailed diagram of the memory reading and writing process.

\section{Implementation Details}
We adopt an Alex-like CNN as in SiamFC \cite{Bertinetto2016} for feature extraction, where the input image sizes of the object  and search images are $127\times 127 \times 3$ and $255 \times 255 \times 3$ respectively. \ty{We use the same strategy for cropping search and object images as in \cite{Bertinetto2016}, where some context margins around the target are added when cropping the object image.} The whole network is trained offline on the VID dataset (object detection from video) of ILSVRC \cite{ILSVRC15} from scratch, and takes about a day. 
Adam \cite{kingma2014adam} optimization is used with a mini-batches of 8 video clips of length 16. The initial learning rate is 1e-4 and is multiplied by 0.8 every 10k iterations. The video clip is constructed by 
uniformly sampling frames \abc{(keeping the temporal order)} from each video. \ytyy{This aims to diversify the appearance variations in one episode for training, which can simulate fast motion, fast background change, jittering object, low frame rate.}
We use data augmentation, including small image stretch and translation for the target image and search image. 
The dimension of memory states in the LSTM controller is 512 and the retain probability used in dropout for LSTM is 0.8. The number of memory slots is $N=8$. The decay factor used for calculating the access vector is $\lambda=0.99$.
%
At test time, the tracker runs completely feed-forward and no online fine-tuning is needed. We locate the target based on the upsampled response map as in SiamFC \cite{Bertinetto2016}, and handle the scale changes by searching for the target over three scales $1.05^{[-1,0,1]}$. \tyy{To smoothen scale estimation and penalize large displacements, we update the object scale with the new one by exponential smoothing $s_{t} = (1-\gamma)*s_{t-1}+\gamma s_{new}$, where $s$ is the scale value and the exponential factor $\gamma = 0.6$. Similarly, we dampen the response map with a cosine window by an exponential factor of 0.15.}

Our algorithm is implemented in Python with the TensorFlow toolbox \cite{abadi2016tensorflow}. It runs at about 50 fps on a computer with four Intel(R) Core(TM) i7-7700 CPU @ 3.60GHz and a single NVIDIA GTX 1080 Ti with 11GB RAM.

\section{Experiments}

We evaluate our proposed tracker, denoted as MemTrack, on three challenging datasets: OTB-2013 \cite{Wu2013}, OTB-2015 \cite{Wu2015} and VOT-2016 \cite{Kristan2016}.  We follow the standard protocols, and evaluate using precision and success plots, as well as area-under-the-curve (AUC).

\subsection{Ablation Studies}\label{abla}

\begin{figure}[t]
	\begin{center}
		\includegraphics[width=0.85\linewidth]{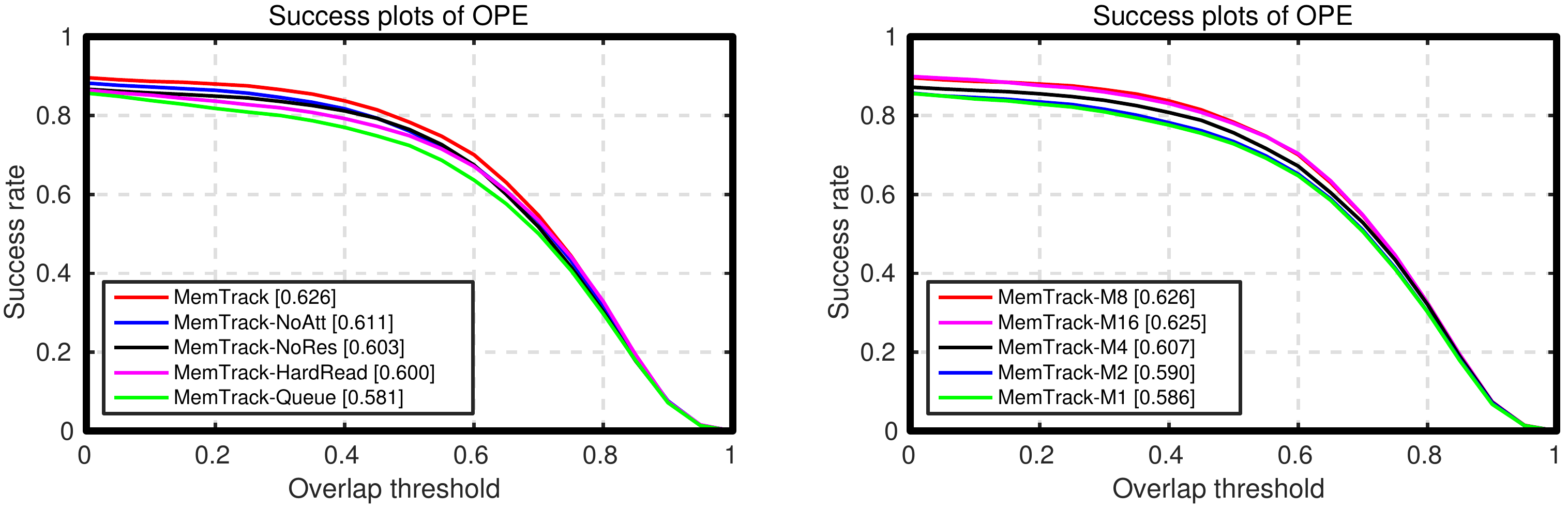}
	\end{center}
	\caption{Ablation studies: (left) success plots of different variants of our tracker on OTB-2015; (right) success plots for different memory sizes \{1, 2, 4, 8, 16\} on OTB-2015. 
	}
	\label{fig:7}
\end{figure}

Our \abc{MemTrack} tracker contains \abc{three important components:} 1) an attention mechanism, which calculates the attended feature vector for memory reading; 2) a dynamic memory network, which maintains the target's appearance variations; and 3) residual template learning, which controls the amount of model updating \abc{for each channel of the template}. To evaluate their separate contributions to our tracker, we implement several variants of our method and verify them on OTB-2015 dataset. 

\yty{ We first design a variant of MemTrack without attention mechanism (MemTrack-NoAtt), which averages all $L$ feature vectors to get the 
	feature vector $\mathbf{a}_t$ \abcn{for the LSTM input.} 
	Mathematically, it changes 
	(2) to $\mathbf{a}_t = \frac{1}{L}\sum_{i=1}^{L}\mathbf{f}^*_{t,i} $. As we can see in Figure \ref{fig:7} (left), Memtrack without attention decreases performance, \abc{which shows the benefit of using attention to roughly localize the target in the search image.}} 
We also design a naive strategy that simply writes the new target template sequentially into the memory slots as a queue (MemTrack-Queue). When the memory is fully occupied, the oldest template will be replaced with the new  template. The retrieved template is generated by averaging all templates stored in the memory slots. As seen in Fig.~\ref{fig:7} (left), such simple approach cannot produce good performance, \abc{which shows the necessity of our dynamic memory network}. \ty{We next devise a hard template reading scheme (MemTrack-HardRead), i.e., retrieving a single template by max cosine distance, to replace the soft weighted sum reading scheme. Figure \ref{fig:7} (left) shows that hard-templates decrease performance possibly due to its non-differentiability }
\yty{To verify the effectiveness of \abc{gated} residual template learning, we design another variant of MemTrack--- removing channel-wise residual gates (MemTrack-NoRes), \emph{i.e.} directly adding the retrieved and initial templates to get the final template. From Fig.~\ref{fig:7} (left), our \abc{gated} residual template learning mechanism boosts the performance as it helps to select correct residual channel features for template updating.}

We also investigate the effect of memory size  on tracking performance. Figure \ref{fig:7} (right) shows success plots on OTB-2015 using different numbers of memory slots. Tracking accuracy increases along with the memory size and saturates at 8 memory slots. Considering the runtime and memory usage, we choose 8 as the default number. 

\subsection{Comparison Results}

We compare our method MemTrack with 9 recent {\em real-time} trackers ($\geq$ 15 fps), including CFNet \cite{Valmadre2017}, LMCF \cite{Wang2017}, ACFN \cite{Choi2017}, RFL \cite{Yang2017}, SiamFC \cite{Bertinetto2016}, SiamFC\_U \cite{Valmadre2017}, Staple \cite{Bertinetto2016-1}, DSST \cite{Danelljan2014}, and KCF \cite{Henriques2015} on both OTB-2013 and OTB-2015. 
To further show our tracking accuracy, we also compared with another 8 recent state-of-the art trackers that are {\em not} real-time speed, including CREST \cite{Song2017},  CSR-DCF \cite{Lukezic2017}, MCPF \cite{Zhang2017}, SRDCFdecon \cite{Danelljan2016}, SINT \cite{Tao2016}, SRDCF \cite{Danelljan2015}, HDT \cite{Qi2016}, HCF \cite{Ma2015} on OTB-2015.

\begin{figure}[t]
	\begin{center}
		\includegraphics[width=0.85\linewidth]{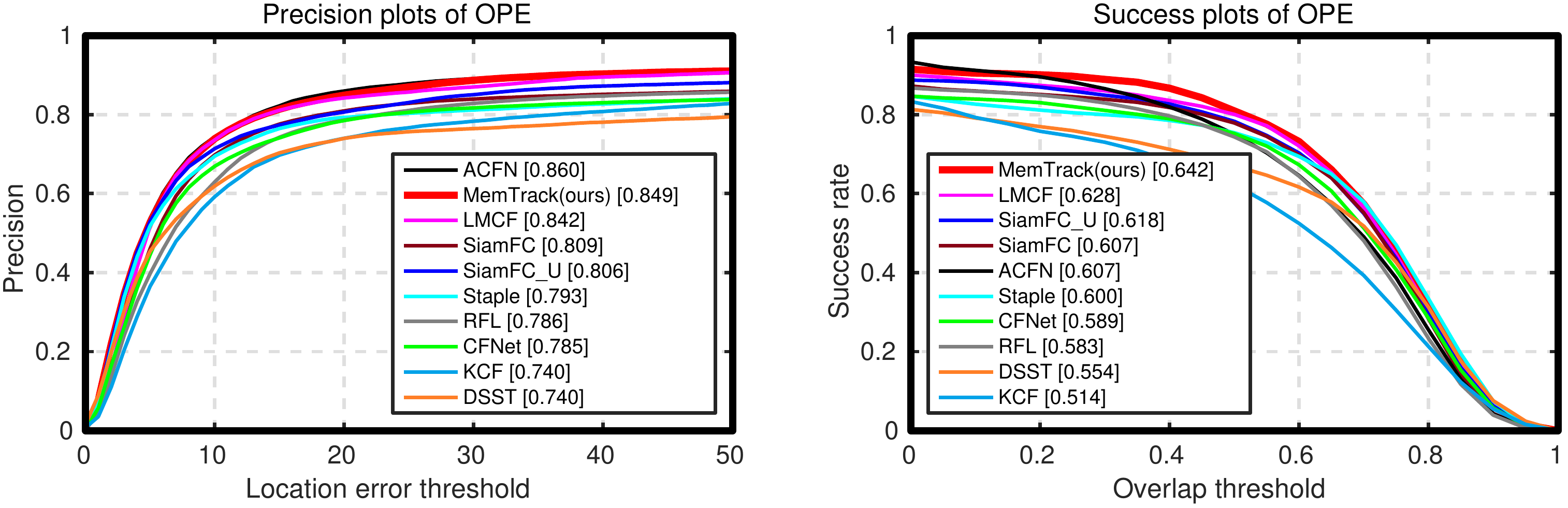}
	\end{center}
	\caption{Precision and success plot on OTB-2013 for recent real-time trackers.
	}
	\label{fig:8}
\end{figure}

\textbf{OTB-2013 Results:} OTB-2013 \cite{Wu2013} dataset contains 51 sequences with 11 video attributes and two evaluation metrics, which are center location error and overlap ratio. Figure \ref{fig:8} shows the one-pass comparison results with recent real-time trackers on OTB-2013. Our tracker achieves the best AUC on the success plot and second place on precision plot. Compared with SiamFC \cite{Bertinetto2016}, which is the baseline for matching-based methods without online updating, our tracker 
achieves an improvement of 4.9\% on precision plot and 5.8\% on success plot.
Our method also outperforms SiamFC\_U, the improved version of SiamFC \cite{Valmadre2017} that uses simple linear interpolation of the old and new filters with a small learning rate for online updating. 
This indicates that our dynamic memory networks can handle object appearance changes better than simply interpolating new templates with old ones.

\begin{figure}[t]
	\begin{center}
		\includegraphics[width=0.85\linewidth]{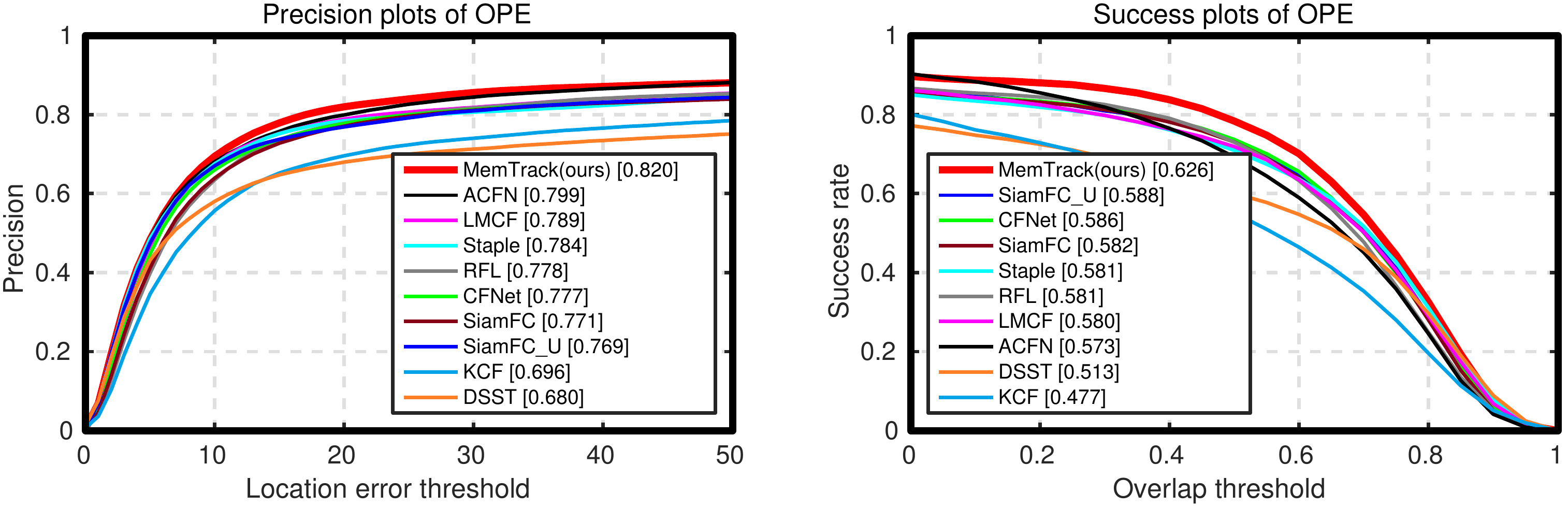}
	\end{center}
	\caption{Precision and success plot on OTB-2015 for recent real-time trackers.}
	\label{fig:9}
\end{figure}

\begin{figure*}[t]
	\begin{center}
		\includegraphics[width=0.95\linewidth]{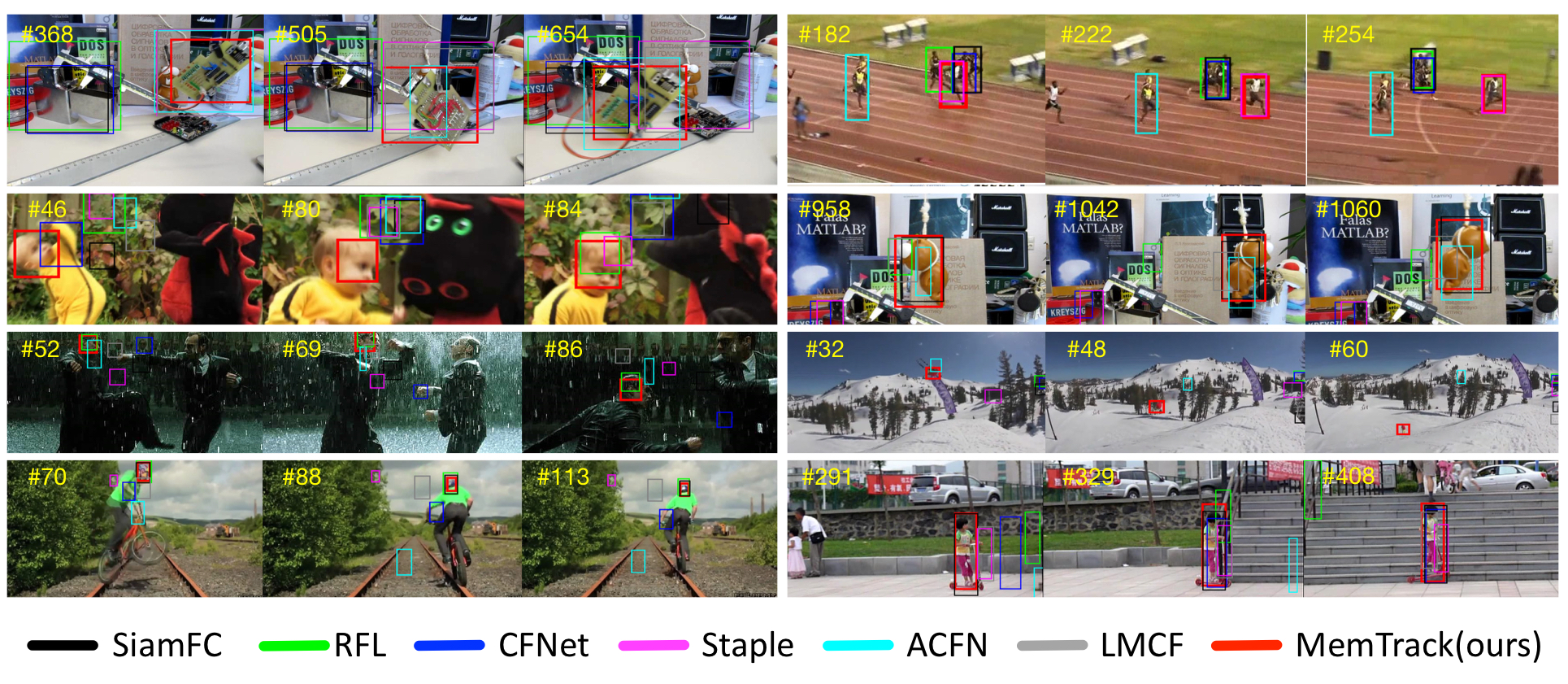}
	\end{center}
	\caption{Qualitative results of our MemTrack, along with  SiamFC \cite{Bertinetto2016}, RFL \cite{Yang2017}, CFNet \cite{Valmadre2017},  Staple \cite{Bertinetto2016-1}, LMCF \cite{Wang2017}, ACFN \cite{Choi2017} on eight challenge sequences. From left to right, top to bottom: \textit{board, bolt2, dragonbaby, lemming, matrix, skiing, biker, girl2}.}
	\label{fig:12}
\end{figure*}

\textbf{OTB-2015 Results:} The OTB-2015 \cite{Wu2015} dataset is the extension of OTB-2013 to 100 sequences, and is thus more challenging.
Figure \ref{fig:9} presents the precision plot and success plot for recent real-time trackers. Our tracker outperforms all other methods in both measures. Specifically, our method performs much better than RFL \cite{Yang2017}, which uses the memory states of LSTM to maintain the object appearance variations. This demonstrates the effectiveness of using an external addressable memory to manage object appearance changes, compared with using LSTM memory which is limited by the size of the hidden states.
Furthermore, MemTrack improves the baseline of template-based method SiamFC \cite{Bertinetto2016} with 6.4\% on precision plot and 7.6\% on success plot respectively. 
Our tracker also outperforms the most recently proposed two trackers, LMCF \cite{Wang2017} and ACFN \cite{Choi2017}, on AUC score with a large margin.
\begin{figure}[t]
	\begin{center}
		\includegraphics[width=0.85\linewidth]{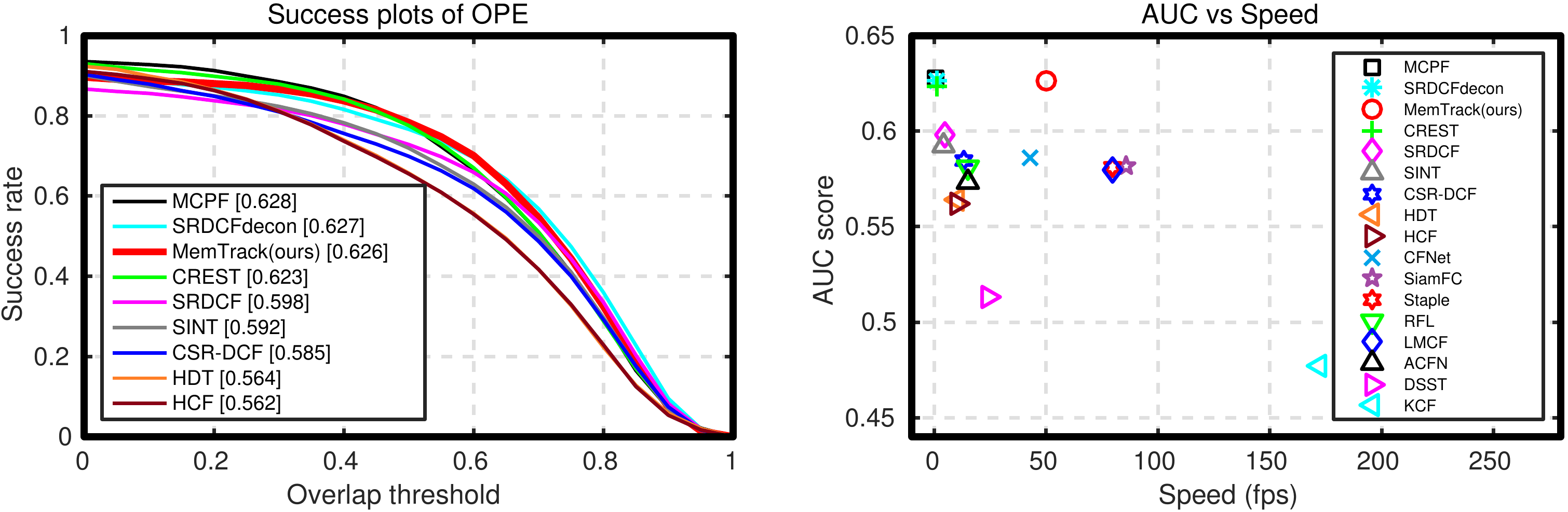}
	\end{center}
	\caption{(left) Success plot on OTB-2015 comparing our real-time MemTrack with recent {\em non-real-time} trackers. (right) AUC score vs speed with recent trackers.}
	\label{fig:10}
\end{figure}
Figure \ref{fig:10} presents the comparison results of 8 recent state-of-the-art {\em non-real time} trackers for AUC score (left plot), and the AUC score vs speed (right plot) of all trackers.
Our MemTrack, which runs in real-time, has similar AUC performance to CREST \cite{Song2017}, MCPF \cite{Zhang2017} and SRDCFdecon \cite{Danelljan2016}, which all run at about 1 fps.
Moreover, our MemTrack also surpasses SINT, which is another matching-based method with optical flow as motion information, in terms of both accuracy and speed.
\begin{figure*}[t]
	\begin{center}
		\includegraphics[width=\linewidth]{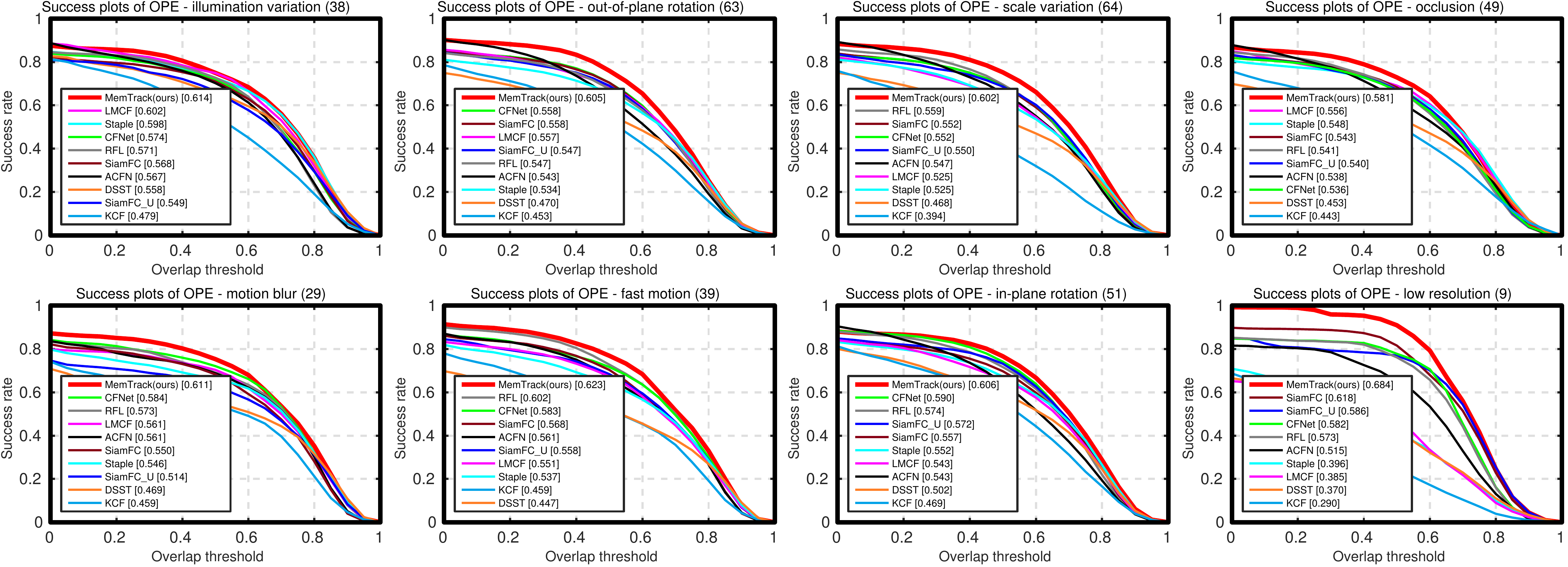}
	\end{center}
	\caption{The success plot of OTB-2015 on eight challenging attributes: illumination variation, out-of-plane rotation, scale variation, occlusion, motion blur, fast motion, in-plane rotation and low resolution }
	\label{fig:11}
\end{figure*}
Figure \ref{fig:11} further shows the AUC scores of real-time trackers on OTB-2015 under different video attributes including illumination variation, out-of-plane rotation, scale variation, occlusion, motion blur, fast motion, in-plane rotation, and low resolution. Our tracker outperforms all other trackers on these attributes. In particular, for the low-resolution attribute, our MemTrack surpasses the second place (SiamFC) with a 10.7\% improvement on AUC score. 
In addition, our tracker also works well under out-of-plane rotation and scale variation.
Fig.~\ref{fig:12} shows some qualitative results of our tracker compared with 6 real-time trackers. 


\begin{table*}
	\small
	\begin{center}
		\begin{tabular}{cccccc|ccccc}
			\hline 
			Trackers & MemTrack & SiamFC & RFL & HCF& KCF & CCOT &TCNN & DeepSRDCF & MDNet \\
			\hline
			EAO ($\uparrow$) & 0.2729 & 0.2352 & 0.2230 &0.2203 & 0.1924&  0.3310 & 0.3249 & 0.2763 &0.2572\\
			A ($\uparrow$) & 0.53 & 0.53  &0.52 &0.44 & 0.48 &  0.54 & 0.55 &0.52  & 0.54\\
			R ($\downarrow$) & 1.44 & 1.91 &2.51 &1.45 &1.95 &  0.89 & 0.83 & 1.23 & 0.91\\
			fps ($\uparrow$) & 50 & 86 & 15& 11& 172&  0.3 & 1 & 1 & 1 \\
			\hline
		\end{tabular} 
	\end{center}
	\caption{Comparison results on VOT-2016 with top performers. The evaluation metrics include expected average overlap (EAO), accuracy and robustness value (A and R), accuracy and robustness rank (Ar and Rr). Best results are bolded, and second best is underlined. The up arrows indicate higher values are better for that metric, while down arrows mean lower values are better.}
	\label{tb:2}
\end{table*}
	

\textbf{VOT-2016 Results:} The VOT-2016 dataset contains 60 video sequences with per-frame annotated visual attributes. Objects are marked with rotated bounding boxes to better fit their shapes. \ty{We compare our tracker with 8 trackers (four real-time and four top-performing)on the benchmark, including SiamFC \cite{Bertinetto2016}, RFL \cite{Yang2017},   HCF \cite{Ma2015}, KCF \cite{Henriques2015},  CCOT \cite{Danelljan2016-1}, TCNN \cite{Nam2016-1}, DeepSRDCF \cite{Danelljan2016-2}, and MDNet \cite{Nam2016}.
Table \ref{tb:2} summarizes results. Although our MemTrack performs worse than \tyy{CCOT, TCNN and DeepSRDCF over EAO}, it runs at 50 fps while others runs at 1 fps or below. Our tracker consistently outperforms the baseline SiamFC and RFL, as well as other real-time trackers.} As reported
in VOT2016, the SOTA bound is EAO 0.251, which
MemTrack exceeds (0.273).

\section{Conclusion}
In this paper, we propose a dynamic memory network with an external addressable memory block for visual tracking, aiming to adapt matching templates to object appearance variations. 
An LSTM with attention scheme controls the memory access by parameterizing the memory interactions. We develop \abc{channel-wise gated} residual template learning to form the final matching model, which preserves the conservative information present in the initial target, while providing online adapability \abc{of each feature channel}. Once the offline training process is finished, no online fine-tuning 
is needed, 
which leads to real-time speed \abcn{of 50 fps}. Extensive experiments on standard tracking benchmark demonstrates the effectiveness 
of our MemTrack.

\noindent \textbf{Acknowledgments} This work was supported by grants from the Research Grants Council of the Hong Kong Special Administrative Region, China (Project No. [T32-101/15-R] and CityU 11212518), and by a Strategic Research Grant from City University of Hong Kong (Project No. 7004887). We are grateful for the support of NVIDIA Corporation with the donation of the Tesla K40 GPU used for this research.
\clearpage

\bibliographystyle{splncs04}
\bibliography{egbib}
\end{document}